# A Transfer-Learning Approach for Accelerated MRI using Deep Neural Networks

Salman Ul Hassan Dar[1,2], Muzaffer Özbey[1,2], Ahmet Burak Çatlı[1,2], Tolga Çukur[1,2,3]

[1]Department of Electrical and Electronics Engineering, Bilkent University, Ankara, Turkey
[2]National Magnetic Resonance Research Center (UMRAM), Bilkent University, Ankara, Turkey
[3]Neuroscience Program, Sabuncu Brain Research Center, Bilkent University, Ankara, Turkey

*Running title:* A transfer-learning approach for accelerated MRI using deep neural networks

*Address correspondence to:*

Tolga Çukur
Department of Electrical and Electronics Engineering, Room 304
Bilkent University
Ankara, TR-06800, Turkey
TEL: +90 (312) 290-1164
E-MAIL: cukur@ee.bilkent.edu.tr

This work was supported in part by a Marie Curie Actions Career Integration Grant (PCIG13-GA- 2013-618101), by a European Molecular Biology Organization Installation Grant (IG 3028), by a TUBA GEBIP fellowship, and by a BAGEP fellowship awarded to T. Çukur. We also gratefully acknowledge the support of NVIDIA Corporation with the donation of the Titan X Pascal GPU used for this research.

To be submitted to Magnetic Resonance in Medicine.

# Abstract

**Purpose:** Neural networks have received recent interest for reconstruction of undersampled MR acquisitions. Ideally network performance should be optimized by drawing the training and testing data from the same domain. In practice, however, large datasets comprising hundreds of subjects scanned under a common protocol are rare. The goal of this study is to introduce a transfer-learning approach to address the problem of data scarcity in training deep networks for accelerated MRI.

**Methods:** Neural networks were trained on thousands of samples from public datasets of either natural images or brain MR images. The networks were then fine-tuned using only few tens of brain MR images in a distinct testing domain. Domain-transferred networks were compared to networks trained directly in the testing domain. Network performance was evaluated for varying acceleration factors (2-10), number of training samples (0.5-4k) and number of fine-tuning samples (0-100).

**Results:** The proposed approach achieves successful domain transfer between MR images acquired with different contrasts ($T_1$- and $T_2$-weighted images), and between natural and MR images (ImageNet and $T_1$- or $T_2$-weighted images). Networks obtained via transfer-learning using only tens of images in the testing domain achieve nearly identical performance to networks trained directly in the testing domain using thousands of images.

**Conclusion:** The proposed approach might facilitate the use of neural networks for MRI reconstruction without the need for collection of extensive imaging datasets.

## Introduction

The unparalleled soft-tissue contrast in MRI has rendered it a preferred modality in many diagnostic applications, but long scan durations limit its clinical use. Acquisitions can be accelerated by undersampling in k-space, and a tailored reconstruction can be used to recover unacquired data. Because MR images are inherently compressible, a popular framework for accelerated MRI has been compressive sensing (CS) (1,2). CS has offered improvements in scan efficiency in many applications including structural (2), angiographic (3), functional (4), diffusion (5), and parametric imaging (6). Yet the CS framework is not without limitation. First, CS involves nonlinear optimization algorithms that scale poorly with growing data size and hamper clinical workflow. Second, CS commonly assumes that MRI data are sparse in fixed transform domains, such as finite differences or wavelet transforms. Recent studies highlight the need for learning the transform domains specific to each dataset to optimize performance (7). Lastly, CS requires careful parameter tuning (e.g., for regularization) for optimal performance. While several approaches were proposed for data-driven parameter tuning (8,9), these methods can induce further computational burden.

Neural network (NN) architectures that reconstruct images from undersampled data have recently been proposed to address the abovementioned limitations. Improved image quality over traditional CS has readily been demonstrated for several applications including angiographic (10), cardiac (11–13), brain (14–34), abdominal (35–37), and musculoskeletal imaging (38–42). The common approach is to train a network off-line using a relatively large set of fully-sampled MRI data, and then use it for on-line reconstruction of undersampled data. Reconstructions can be achieved in several hundred milliseconds, significantly reducing computational burden (39,43). The NN framework also alleviates the need for adhoc selection of transform domains. For example, a recent study used a cascade of CNNs to recover images directly from zero-filled Fourier reconstructions of undersampled data (11,16,41). The trained CNN layers reflect suitable transforms for image reconstruction. The NN framework introduces more tunable hyperparameters (e.g., number of layers, units, activation functions) than would be required in CS. However, previous studies demonstrate that hyperparameters optimized during the training phase generally perform well in the testing phase (43). Taken together, these advantages render the NN framework a promising avenue for accelerated MRI.

A common strategy to enhance network performance is to boost model complexity by increasing the number of layers and units in the architecture. A large set of training data must then be used to reliably learn the numerous model parameters (44). Previous studies either used an extensive database of MR images comprising several tens to hundreds of subjects (12,18,45), or data augmentation procedures to artificially expand the size of training data (11,12). For instance, an early study performed training on $T_1$-weighted brain images from nearly 500 subjects in the human connectome project (HCP) database, and testing on $T_2$-weighted images (18). Yet, it remains unclear how well a network trained on images acquired with a specific type of tissue contrast generalize to images acquired with different contrasts. Furthermore, for optimal reconstruction performance the network must be trained on images acquired with the same scan protocol that it later will be tested on. However, large databases such as those provided by the HCP may not be readily available in many applications, potentially rendering NN-based reconstructions suboptimal.

In this study, we propose a transfer-learning approach to address the problem of data scarcity in network training for accelerated MRI. In transfer-learning, network training is performed in some domain where large datasets are available, and knowledge captured by the trained network is then transferred to a different domain where data are scarce (46,47). Domain transfer was previously used to suppress coherent aliasing artifacts in projection reconstruction acquisitions (30), to perform non-Cartesian to Cartesian interpolation in k-space (18), and to assess the robustness of network reconstructions to variations in SNR and undersampling patterns (42). In contrast, we employ transfer-learning to enhance NN-based reconstructions of randomly undersampled acquisitions in the testing domain. A deep CNN architecture with multiple subnetworks is taken as a model network (11). For reconstruction of multi-coil data, calibration consistency, data consistency and CNN blocks are incorporated to synthesize missing samples. In the training domain using several thousand images, each subnetwork is trained sequentially to reconstruct reference images from zero-filled reconstructions of undersampled data. The full network is then fine-tuned end-to-end in the testing domain using few tens of images.

To demonstrate the proposed approach, comprehensive evaluations were performed across a broad range of acceleration factors (R=2-10) on $T_1$- and $T_2$-weighted brain images. Separate network models were learnt for domain transfer between MR images acquired with different contrasts ($T_1$- and $T_2$-weighted), and between natural and MR images (ImageNet and $T_1$- or $T_2$-weighted). Domain-transferred networks were quantitatively compared against networks trained in the testing domain, and against conventional CS reconstructions. We find that domain-transferred networks fine-tuned with tens of images achieve nearly identical performance to networks trained directly in the testing domain using thousands of images, and that both networks outperform CS methods.

Note – A preliminary version of this work was presented at the 26th Annual Meeting of ISMRM under the title "Transfer learning for reconstruction of accelerated MRI acquisitions via neural networks" (48).

## Methods

**MRI Reconstruction via Compressed Sensing (CS)**

*Single-coil data.* In accelerated MRI, an undersampled acquisition is followed by a reconstruction to recover missing k-space samples. This recovery can be formulated as a linear inverse problem:

$$F_u x = y_u \qquad (1)$$

where $x$ denotes the image to be reconstructed, $F_u$ is the partial Fourier transform operator at the sampled k-space locations, and $y_u$ denotes acquired k-space data. Since Eq. 1 is underdetermined, additional prior information is typically incorporated in the form of a regularization term:

$$x_{rec} = \min_x \|F_u x - y_u\|_2 + R(x) \qquad (2)$$

Here, the first term enforces consistency between acquired and reconstructed data, whereas $R(x)$ enforces prior information to improve reconstruction performance. In CS, $R(x)$ typically corresponds to L$_1$-norm of the image in a known transform domain (e.g., wavelet transform or finite differences transform).

The solution of Eq. 2 involves non-linear optimization algorithms that are often computationally complex. This reduces clinical feasibility as reconstructions times become prohibitive with increasing size of data. Furthermore, assuming ad hoc selection of fixed transform domains leads to suboptimal reconstructions in many applications (7). Lastly, it is often challenging to find a set of reconstruction parameters that work optimally across subjects (49).

*Multi-coil data.* For reconstruction of multi-coil data, a hybrid parallel imaging/compressed sensing approach is commonly used, such as SPIRiT (iTerative Self-consistent Parallel Imaging Reconstruction). In SPIRiT (50), the recovery problem in Eq. 2 can be reformulated as:

$$x_{rec} = \min_x \|F_u x - y_u\|_2 + \|(G - I)Fx\|_2 + R(x) \qquad (3)$$

where $x$ denotes multi-coil images to be reconstructed, $y_u$ denotes acquired multi-coil k-space data, $F$ is the forward Fourier transform operator and $G$ denotes the interpolation operator that synthesizes unacquired samples in terms of acquired samples across neighboring k-space and coils. To enforce sparsity, $R(x)$ can be selected as the L$_1$-norm of wavelet coefficients. One efficient way to solve Eq. 3 is via the projection onto convex sets (POCS) algorithm (51). POCS alternates among a calibration-consistency (CC) projection that applies G, a sparsity projection that enforces sparsity in the transform domain, and a data-consistency (DC) projection.

**MRI Reconstruction via Neural Networks (NN)**

*Single-coil data.* In the NN framework, a network architecture is used for reconstruction instead of explicit transform-domain constraints. Network training is performed via a supervised learning procedure, with the aim to find the set of network parameters that yield accurate reconstructions undersampled acquisitions. This procedure is performed on a large set of training data (with $N_{train}$ samples), where fully-sampled reference acquisitions are retrospectively undersampled. Network training typically amounts to minimizing the following loss function (29):

$$\min_{\theta} \sum_{n=1}^{N_{train}} \frac{1}{N_{train}} \|C(x_{un}; \theta) - x_{refn}\|_2 \quad (4)$$

where $x_{un}$ represents the Fourier reconstruction of $n^{th}$ undersampled acquisition, $x_{refn}$ represents the respective Fourier reconstruction of the fully-sampled acquisition, $C(x_{un}; \theta)$ denotes the output of the network given the input image $x_{un}$ and the network parameters $\theta$. To reduce sensitivity to outliers, here we minimized a hybrid loss that includes both mean-squared error and mean-absolute error terms. To minimize over-fitting, we further added an L2-regularization term on the network parameters. Therefore, neural network training was performed with the following loss function:

$$\min_{\theta} \sum_{n=1}^{N_{train}} \frac{1}{N_{train}} \|C(x_{un}; \theta) - x_{refn}\|_2 + \sum_{n=1}^{N_{train}} \frac{1}{N_{train}} \|C(x_{un}; \theta) - x_{refn}\|_1 + \gamma_\Phi \|\theta\|_2 \quad (5)$$

where $\gamma_\Phi$ is the regularization parameter for network parameters.

A network trained on a sufficiently large set of training examples can then be used to reconstruct an undersampled acquisition from an independent test dataset. This reconstruction can be achieved by reformulating the problem in Eq. 2 (29):

$$x_{rec} = \min_{x} \lambda \|F_u x - y_u\|_2 + \|C(x_u; \theta^*) - x\|_2 \quad (6)$$

where $C(x_u; \theta^*)$ is the output of the trained network with optimized parameters $\theta^*$. Note that the problem in Eq. 6 has the following closed-form solution (29):

$$y_{rec}(k) = \begin{cases} \dfrac{[FC(x_u; \theta^*)](k) + \lambda y_u(k)}{1 + \lambda}, & \text{if } k \in \Omega \\ [FC(x_u; \theta^*)](k), & \text{otherwise} \end{cases} \quad (7)$$

$$x_{rec} = F^{-1} y_{rec}$$

where $k$ denotes k-space location, $\Omega$ represents the set of acquired k-space locations, $F$ and $F^{-1}$ are the forward and backward Fourier transform operators, and $x_{rec}$ is the reconstructed image. The solution outlined in Eq. 7 performs two separate projections during reconstruction. The first projection calculates the output of the trained neural network $C(x_u; \theta^*)$ given the input image $x_u$, the Fourier reconstruction of undersampled data. The second projection enforces data consistency. The parameter $\lambda$ in Eq. 7 controls the relative weighing between data samples that are originally acquired and those that are recovered by the network. Here we used $\lambda = \infty$ to enforce data consistency strictly. The projection outlined in Eq. 7 can be compactly expressed as (11):

$$f_{DC}\{C(x_u; \theta^*)\} = F^{-1} \Lambda F C(x_u; \theta^*) + \frac{\lambda}{1 + \lambda} F^{-1} x_u \quad (8)$$

where $\Lambda$ is a diagonal matrix:

$$\Lambda_{kk} = \begin{cases} \frac{1}{1+\lambda}, & if\ k\ \epsilon\ \Omega \\ 1, & otherwise \end{cases} \quad (9)$$

Conventional optimization algorithms for CS run iteratively to progressively minimize the loss function. A similar approach can also be adopted for NN-based reconstructions (11,16,41). Here, we cascaded several subnetworks in series with DC projections interleaved between consecutive subnetworks (11). In this architecture, the input $x_{ip}$ to the $p^{th}$ subnetwork was formed as:

$$x_{ip} = \begin{cases} x_{un}, & if\ p = 1 \\ f_{DC}\{C_{p-1}(f_{DC}\{C_{p-2}(f_{DC}\ ....\ C_1(x_{un};\theta_1^*)\};\theta_{p-1}^*)\}, & if\ p > 1 \end{cases} \quad (10)$$

where $\theta_p^*$ denotes the parameters of the $p^{th}$ subnetwork. Starting with the initial network with $p = 1$, each subnetwork was trained sequentially by solving the following optimization problem:

$$\min_{\theta_p} \sum_{n=1}^{N_{train}} \frac{1}{N_{train}} \|C(x_{ip};\theta_p) - x_{refn}\|_2 + \sum_{n=1}^{N_{train}} \frac{1}{N_{train}} \|C(x_{ip};\theta_p) - x_{refn}\|_1 + \gamma_\Phi \|\theta_p\|^2 \quad (11)$$

While training the $p^{th}$ subnetwork, the parameters of preceding networks and thus the input $x_{ip}$ are assumed to be fixed.

*Multi-coil data.* Similar to SPIRiT, for multi-coil reconstructions, here we reformulate Eq. 3 as:

$$x_{rec} = \min_x \|F_u x - y_u\|_2 + \|(G - I)Fx\|_2 + \|C(A^* x_u;\theta^*) - A^* x\|_2 \quad (12)$$

where $x$ denotes the multi-coil images to be reconstructed, $A$ and $A^*$ denote coil-sensitivity profiles and adjoints, and G denotes the interpolation operator in SPIRiT as in Eq. 3. The network C has been trained to recover fully-sampled coil-combined images given undersampled coil-combined images as outlined in Eq. 5. The trained network regularizes the reconstruction in Eq. 12 given undersampled coil-combined images $A^* x_u$. The optimization problem in Eq. 12 is solved by alternating projections for calibration-consistency (CC), data-consistency (DC) and neural-network (CNN) consistency. Subnetworks are cascaded in series with data consistency and calibration consistency projections. The data-consistency projection can be compactly expressed as:

$$f_{CC}\{x_u\} = GFx_u \quad (13)$$

In this multi-coil implementation, the input $x_{ip}$ to the $p^{th}$ subnetwork was formed as:

$$x_{ip} = \begin{cases} f_{DC}\{f_{CC}\{x_{un}\}\}, & if\ p = 1 \\ f_{DC}\{f_{CC}\{f_{DC}\{C_{p-1}(f_{DC}\{f_{CC}\{(...f_{DC}\{C_1(f_{DC}\{f_{CC}\{x_{un}\}\};\theta_1^*)\};\theta_{p-1}^*)\}, & if\ p > 1 \end{cases} \quad (14)$$

Note that both calibration-consistency and neural-network consistency are followed by a data-consistency layer.

**Datasets**

*Public datasets.* For demonstrations on single-coil data, three distinct datasets were used: natural, $T_1$-weighted brain, and $T_2$-weighted brain images. The details are listed below.

1) Natural images: We assembled 5000 natural images from the validation set used during the ImageNet Large Scale Visual Recognition Challenge 2011 (ILSVRC2011) (52). 4000 images were used for training and 1000 images were used for validation. All images were either cropped or zero-padded to yield consistent dimensions of 256x256. Color RGB images were first converted to LAB color space, and the L-channel was extracted to obtain grayscale images.

2) $T_1$-weighted images: We assembled a total of 6160 $T_1$-weighted images (52 subjects) from the MIDAS database (53). These images were divided into 4240 training images (36 subjects), 720 fine-tuning images (6 subjects) and 1200 testing images (10 subjects). In the training phase, 4000 images (34 subjects) were used for training while 240 images (2 subjects) were reserved for validation. In the fine-tuning phase, 480 images (4 subjects) were used for fine-tuning and 240 images (2 subjects) were reserved for validation. There was no overlap between subjects included in the training, validation and testing sets. $T_1$-weighted images analyzed here were collected on a 3T scanner via the following parameters: a 3D gradient-echo sequence, TR=14ms, TE=7.7ms, flip angle=$25^0$, matrix size=256x176, 1 mm isotropic resolution.

3) $T_2$-weighted images: We assembled a total of 5800 $T_2$-weighted images (58 subjects) from the MIDAS database (53). These images were divided into 4200 training images (42 subjects), 600 fine-tuning images (6 subjects) and 1000 testing images (10 subjects), with no subject overlap between training, validation and testing sets. In the training phase, 4000 images (40 subjects) were used for training and 200 images (2 subjects) were used for validation. In the fine-tuning phase, 400 images (4 subjects) were used for fine-tuning and 200 images (2 subjects) were used for validation. $T_2$-weighted images analyzed here were collected on a 3T scanner via the following parameters: a 2D spin-echo sequence, TR=7730ms, TE=80ms, flip angle=$90^0$, matrix size=256x192, 1 mm isotropic resolution.

*Multi-coil MR images*. $T_1$-weighted brain images from 10 subjects were acquired. Within each subject, 60 central cross-sections containing sizeable amount of brain tissue were selected. Images were then divided into 360 training images (6 subjects), 60 validation images (1 subject) and 180 testing images (3 subjects), with no subject overlap. Images were collected on a 3T Siemens Magnetom scanner (maximum gradient strength of 45mT/m and slew rate of 200 T/m/s) using a 32-channel receive-only head coil. The protocol parameters were: a 3D MP-RAGE sequence, TR=2000ms, TE=5.53ms, flip angle=$20^0$, matrix size=256x192x80, 1 mm isotropic resolution. Imaging protocol was approved by the local ethics committee at Bilkent University and all participants provided written informed consent. To reduce computational complexity, geometric-decomposition coil compression (GCC) was performed to reduce number of coils from 32 to 8 (54).

*Multi-coil natural images*. To perform domain transfer from natural images to multi-coil MRI, complex natural images were simulated from 2000 magnitude images in ImageNet by adding sinusoidal phase at random spatial frequencies along each axis varying from $-\pi$ to $+\pi$. The amplitude of the sinusoids was normalized between 0 and 1. Fully-sampled multi-coil $T_1$-weighted acquisitions from 2 training subjects were selected to extract coil-sensitivity maps using ESPIRiT (55). Each multi-coil complex natural image was then simulated by utilizing coil-sensitivity maps of a randomly selected cross-section from the 2 reserved subjects (see Supp. Figure 1 for sample multi-coil complex natural images).

*Undersampling patterns*. Images in each dataset were undersampled via variable-density Poisson-disc sampling (50). All datasets were undersampled for varying acceleration factors (R= 2, 4, 6, 8, 10). Fully sampled images were first Fourier transformed and then retrospectively undersampled. To ensure reliability against mask selection, 100 unique undersampling masks were generated and used during the training phase. A different set of 100 undersampling masks were used during the testing phase.

**Network training**

We adopted a cascade of neural networks as inspired by (11). Five subnetworks were cascaded in series. Each subnetwork contained an input layer, four convolutional layers and an output layer. The input layer consisted of two channels for real imaginary parts of undersampled images. Each convolution operation in the convolutional layers was passed through a rectified linear unit (ReLU) activation. The hidden layers consisted of 64 channels. For single-coil data, the output layer consisted of only a single channel for a magnitude reconstruction. For multi-coil data, separate subnetworks were trained with real and imaginary parts of the reconstruction.

*Subnetwork training*. Subnetworks were trained via the back-propagation algorithm (56). In the forward passes, a batch of 50 samples were passed through the network to calculate the respective loss function. In the backward passes, network parameters were updated according to the gradients of this function with respect to the parameters. The gradient of the loss function with respect to parameters of the $m^{th}$ hidden layer ($\theta_m$) can be calculated using chain rule:

$$\frac{\partial L}{\partial \theta_m} = \frac{\partial L}{\partial o_l}\frac{\partial o_l}{\partial a_l}\frac{\partial a_l}{\partial \theta_l}\frac{\partial o_{l-1}}{\partial a_{l-1}}\cdots\frac{\partial o_m}{\partial a_m}\frac{\partial a_m}{\partial \theta_m} \quad (15)$$

where $l$ is the output layer of the network, $a_l$ is the output of the $l^{th}$ layer, and $o_l$ is the output of the $l^{th}$ layer passed through the activation function. The parameters of the $m^{th}$ layer were only updated if the loss-function gradient flows through all subsequent layers (is non-zero). Each subnetwork was trained individually for 20 epochs. In the training phase, the network parameters were optimized using the ADAM optimizer with a learning rate of $\eta=10^{-4}$, decay rate for first moment of gradient estimates of $\beta_1= 0.9$ and decay rate for the second moment of gradient estimate of $\beta_2=0.999$ (57). Connection weights were L$_2$-regularized with a regularization parameter of $\gamma_\Phi=10^{-6}$.

*Fine tuning*. Networks formed by sequential training of the subnetworks were fine tuned. Here, end-to-end fine tuning was performed on the entire neural-network architecture. To do this, the gradients must be calculated through CNN, DC and CC blocks. The gradient flow through the convolutional subnetworks that contain basic arithmetic operations and ReLU activation functions are well known. The gradient flow through DC in Eq. 7 with respect to its input $C(A^*x_u;\theta^*)$ is given as:

$$\frac{\partial f_{DC}}{\partial C(A^*x_u;\theta^*)} = F^{-1}\Lambda F \quad (16)$$

due to linearity of the Fourier operator ($F$). Similarly, the gradient flow through CC in Eq. 12 with respect to output of the preceding subnetwork $C(A^*x_u;\theta^*)$ is given as:

$$\frac{\partial f_{CC}}{\partial C(A^*x_u;\theta^*)} = \frac{\partial f_{CC}}{\partial f_{DC}}\frac{\partial f_{DC}}{\partial C(A^*x_u;\theta^*)} = GFF^{-1}\Lambda F \quad (17)$$

During the fine-tuning phase, the ADAM optimizer was used with identical parameters to those used in subnetwork training, apart from a lower learning rate of $10^{-5}$ and a total of 100 epochs.

**Performance analyses**

*Single-coil data.* We first evaluated the performance of networks under implicit domain transfer (i.e., without fine tuning). We reasoned that a network trained and tested in the same domain should outperform networks trained and tested on different domains. To investigate this issue, we reconstructed undersampled $T_1$-weighted acquisitions using the ImageNet-trained and $T_2$-trained networks for varying acceleration factors (R=2, 4, 6, 8, 10). The reconstructions obtained via these two networks were compared with reference reconstructions obtained from the network trained directly on $T_1$-weighted images. To ensure that our results were not biased by the selection of a specific MR contrast as the test set, we also reconstructed undersampled $T_2$-weighted acquisitions using the ImageNet-trained and $T_1$-trained networks. The reconstructions obtained via these two networks were compared with reference reconstructions obtained from the network trained directly on $T_2$-weighted images.

Next, we evaluated the performance of network under explicit domain transfer (i.e., with fine tuning). Networks were fine-tuned end-to-end in the testing domain. When $T_1$-weighted images were the testing domain, ImageNet-trained and $T_2$-trained networks were fine-tuned using a small set of $T_1$-weighted images with size ranging in [0 100]. When $T_2$-weighted images were the testing domain, ImageNet-trained and $T_1$-trained networks were fine-tuned using a small set of $T_2$-weighted images with size ranging in [0 100]. In both cases, the performance of fine-tuned networks was compared with the networks trained and further fine-tuned end-to-end directly in the testing domain on 100 images.

Reconstruction performance of a fine-tuned network likely depends on the number of both training and fine-tuning images. To examine potential interaction between the number of training and fine-tuning samples, separate networks were trained using training sets of varying size in [500 4000]. Each network was then fine-tuned using sets of varying size in [0 100]. Performance was evaluated to determine the number of fine-tuning samples that are required to achieve near-optimal performance for each separate size of training set. Optimal performance was taken as the peak signal-to-noise ratio (PSNR) of a network trained directly in the testing domain.

NN-based reconstructions were also compared to those obtained by conventional CS (2). Single-coil CS reconstructions were implemented via a nonlinear conjugate gradient method. Daubechies-4 wavelets were selected as the sparsifying transform. Parameter selection was performed to maximize PSNR on the validation images from the fine-tuning set. Consequently, an $L_1$-regularization parameter of $10^{-3}$, and 80 iterations for $T_1$-weighted acquisitions and 120 iterations for -weighted acquisitions were observed to yield near-optimal performance broadly across R.

*Multi-coil data.* We also demonstrated the proposed approach on multi-coil MR images. For this purpose, a network was trained on 2000 synthetic multi-coil complex natural images (see Methods for details). The network was then fine-tuned using a set of multi-coil $T_1$-weighted images with varying size in [0 100]. This set was randomly selected from the training subjects. Reconstruction performance was compared with networks trained using 360 $T_1$-weighted multi-coil MR images (6 subjects) and $L_1$-SPIRiT (50). A POCS implementation of SPIRiT was used. For each R, parameter selection was performed to maximize PSNR on validation images drawn from the multi-coil MR image dataset. An interpolation kernel width of 7, a Tikhonov regularization parameter of $10^{-2}$ for calibration, an $L_1$-regularization parameter of $10^{-3}$ were observed to yield near-optimal performance across R. Meanwhile, the optimal number of iterations varied based on acceleration factor. For R= [2, 4, 6, 8, 10], the following number of iterations= [20, 30, 45, 65, 80] were selected. The interpolation kernels optimized for SPIRiT were used in the calibration-consistency layers of the networks that contained 5 consecutive CC projections.

To quantitatively compared alternative methods, we measured the structural similarity index (SSIM) and peak signal-to-noise ratio (PSNR) between the reconstructed and fully-sampled reference images. For multi-coil data, the reference image was taken as the coil-combined image obtained via weighted linear combination using coil sensitivity maps from ESPIRiT. The training and testing of NN architectures were performed in the TensorFlow framework (58) using 2 NVIDIA Titan X Pascal GPUs (12 GB VRAM). Single-coil CS reconstructions were performed via libraries in the SparseMRI V0.2 toolbox available at https://people.eecs.berkeley.edu/~mlustig/Software.html. Multi-coil CS reconstructions were performed via libraries in the SPIRiT V0.3 toolbox available at https://people.eecs.berkeley.edu/~mlustig/Software.html.

## Results

A network trained on the same type of images that it will later be tested on should outperform networks trained and tested on different types of images. However, this performance difference should diminish following successful domain transfer between the training and testing domains. To test this prediction, we first investigated generalization performance for implicit domain transfer (i.e., without fine tuning) in a single-coil setting. The training domain contained natural images from the ImageNet database or $T_2$-weighted images, and the testing domain contained $T_1$-weighted images. Figure 2 displays reconstructions of an undersampled $T_1$-weighted acquisition via the ImageNet-trained, $T_2$-trained and $T_1$-trained networks for R=4. As expected, the $T_1$-trained network yields sharper and more accurate reconstructions compared to the raw ImageNet–trained and $T_2$-trained networks. Next, we examined explicit domain transfer where ImageNet-trained and $T_2$-trained networks were fine-tuned. In this case, all networks yielded visually similar reconstructions. Furthermore, when compared against conventional compressive sensing (CS), all network models yielded superior performance. Figure 3 displays reconstructions of an undersampled $T_1$-weighted acquisition via the ImageNet-trained, $T_2$-trained and $T_1$-trained networks, and CS for R=4. The ImageNet-trained network produces images of similar visual quality to other networks and it outperforms CS in terms of image sharpness and residual aliasing artifacts.

To corroborate these visual observations, reconstruction performance was quantitively assessed for both implicit and explicit domain transfer across R=2-10. PSNR and SSIM measurements across the test set are listed in Table 1 and Supp. Table 1. For implicit domain transfer, the $T_1$-trained networks outperform domain-transferred networks and CS consistently across all R. For explicit domain transfer, the differences between the $T_1$-trained and domain-transferred networks diminish. Following fine-tuning, the average differences in (PSNR, SSIM) across R between ImageNet and $T_1$-trained networks diminish from (1.97dB, 3.80%) to (0.18dB, 0.20%), and difference between $T_2$-trained and $T_1$-trained networks diminish from (1.34dB, 2.20%) to (0.04dB, 0%). Furthermore, the domain-transferred networks outperform CS consistently across R, by an average of 4.00dB PSNR and 9.9% SSIM.

Next, we repeated the analyses for implicit and explicit domain transfer when the testing domain contained $T_2$-weighted images. Supp. Figure 2 displays reconstructions of an undersampled $T_2$-weighted acquisition via the ImageNet-, $T_1$- and $T_2$-trained networks for acceleration factor R=4. Again, the network trained directly in the testing domain ($T_2$-weighted) outperforms domain transferred networks. After fine tuning with as few as 20 images, the domain-transferred networks yield visually similar reconstructions to the $T_2$-trained network.

PSNR and SSIM measurements on $T_2$-weighted reconstructions across the test set are listed in Supp. Table 2. Following fine-tuning, average (PSNR, SSIM) differences between ImageNet and $T_2$-trained networks diminish from (1.23dB, 3.40%) to (0.19dB, 0.40%), and difference between $T_1$-trained and $T_2$-trained networks diminish from (1.14dB, 2.80%) to (0.14dB, -0.20%). Across R, the domain-transferred networks also outperform CS by 5.21dB PSNR and 12.5% SSIM.

Reconstruction performance of domain-transferred networks may depend on the sizes of both training and fine-tuning sets. To examine interactions between the number of training ($N_{train}$) and fine-tuning ($N_{tune}$) samples, we trained networks using training sets in the range [500 4000] and fine-tuning sets in the range [0 100]. Figure 4 shows average PSNR values for a reference

$T_1$-trained network trained on 4000 and fine-tuned on 100 images, and domain transferred networks for R=2-10. Without fine-tuning, the $T_1$-trained network outperforms both domain-transferred networks. As the number of fine-tuning samples increases, the PSNR differences decay gradually to a negligible level. Consistently across R, domain-transferred networks trained on smaller training sets require more fine-tuning samples to yield similar performance.

Figure 5 displays the number of fine-tuning samples required for the PSNR values for ImageNet-trained networks to converge for R=2-10. Convergence was taken as the number of fine-tuning samples where the percentage change in PSNR by incrementing number of fine-tuning samples fell below 0.05% of PSNR for the $T_1$-trained network. Consistently across R, networks trained on fewer samples require more fine-tuning samples for convergence. However, the required number of fine-tuning samples is greater for higher R. Averaged across R, $N_{tune}$=72 for $N_{train}$=500, $N_{tune}$=57 for $N_{train}$=1000, $N_{tune}$=58 for $N_{train}$=2000, $N_{tune}$=44 for $N_{train}$=4000.

We also examined interactions between the number of training and fine-tuning samples when the target domain contained $T_2$-weighted images. Supp. Figure 4 shows average PSNR values for a reference $T_2$-trained network trained on 4000 and fine-tuned on 100 images, and domain transferred networks for R=2-10. Similar to the case of $T_1$-weighted images, domain-transferred networks trained on smaller sets require more fine-tuning samples to yield comparable performance. Supp. Figure 5 displays the number of fine-tuning samples required for convergenge of ImageNet-trained networks. Averaged across R=2-10, $N_{tune}$=66 for $N_{train}$=500, $N_{tune}$=46 for $N_{train}$=1000, $N_{tune}$=51 for $N_{train}$=2000, $N_{tune}$=43 for $N_{train}$=4000.

Next, we demonstrated the proposed approach on multi-coil $T_1$-weighted images. We compared ImageNet- and $T_1$-trained networks at R=2-10. Figure 6 displays average PSNR values for the $T_1$-trained network (trained and fine-tuned on 360 images) and ImageNet-trained network (trained on 2000 multi-coil natural images and fine-tuned on [0,100] $T_1$-weighted images). As $N_{tune}$ increases, the PSNR differences between $T_1$- and ImageNet-trained networks start diminishing. Figure 7 displays the number of fine-tuning samples required for the PSNR values for ImageNet-trained networks to converge. Averaged across R=2-10, ImageNet-trained networks require $N_{tune}$=31 for convergence. We also compared the proposed transfer learning approach with $L_1$-regularized SPIRiT. Figure 8 shows representative reconstructions obtained via the ImageNet-trained network, $T_1$-trained network and SPIRiT for R=10. The ImageNet-trained network produces images of similar visual quality to the $T_1$-trained network, and it outperforms SPIRiT in terms of residual aliasing artifacts.

Quantitative assessment of multi-coil reconstructions for the ImageNet-trained network, $T_1$-trained network and SPIRiT across R=2-10 are listed in Table 2. For implicit domain transfer, the $T_1$-trained network performs better than the ImageNet-trained network. Following fine-tuning, the average differences in (PSNR, SSIM) across R between ImageNet and $T_1$-trained networks diminish from (1.92dB, 1.00%) to (0.56dB, 0.10%). Furthermore, the ImageNet-trained network outperforms SPIRiT in all cases, except for R=2 where SPIRiT yields higher PSNR and SSIM, and R=4 where the two methods yield similar PSNR. On average across R, the ImageNet-trained network improves performance over SPIRiT by 0.67dB PSNR and 0.50% SSIM.

**Discussion**

Neural networks for MRI reconstruction involve many free parameters to be learnt, so an extensive amount of training samples is typically needed (59). In theory, network performance should be optimized by drawing the training and testing samples from the same domain, acquired under a common MRI protocol. In practice, however, compiling large public datasets can require coordinated efforts among multiple imaging centers, and so such datasets are rare. As an alternative, several recent studies trained neural networks on a collection of multi-contrast images (19). When needed, data augmentation procedures were used to further expand the training dataset (11,12). While these approaches gather more samples for training, it remains unclear how well a network trained on images acquired with a specific type of tissue contrast generalizes to images acquired with different contrasts. Thus, variability in MR contrasts can lead to suboptimal reconstruction performance.

Here, we first questioned the generalizability of neural network models across different contrasts. We find that a network trained on MR images of a given contrast (e.g. $T_1$-weighted) yields suboptimal reconstructions on images of a different contrast (e.g. $T_2$-weighted). This confirms that the best strategy is to train and test networks in the same domain. Yet, it may not be always feasible to gather a large collection of images from a desired contrast. To address the problem of data scarcity, we proposed a transfer-learning approach for accelerated MRI. The proposed approach trains neural networks using training samples from a large public dataset of natural images. The network is then fine-tuned end-to-end using only few tens of MR images. Reconstructions obtained via the ImageNet-trained network are of nearly identical quality to reconstructions obtained by networks trained directly in the testing domain using thousands of MR images.

Several recent studies have considered domain transfer to enhance performance in NN-based MRI reconstruction (18,29,30,42). A group of studies have aimed to perform implicit domain transfer across MRI contrasts without fine-tuning. One proposed method was to train networks on MR images in a given contrast, and then to directly use the trained networks on images of different contrasts (18). While this method yields successful reconstructions, our results suggest that network performance can be further boosted with additional fine-tuning in the testing domain. Another method to enhance generalizability was to compound datasets containing a mixture of distinct MRI contrasts during network training (29). This approach enforces the network to better adapt to variations in tissue contrast. Yet, in the absence of contrast-specific fine-tuning, networks may deliver suboptimal performance for some individual contrasts.

A second group of studies have attempted explicit domain transfer across training and testing domains via fine-tuning. A recent proposed method trained a deep residual network to remove streaking artifacts from CT images, and the trained network was then used to suppress aliasing artifacts in projection-reconstruction MRI (30). This method leverages the notion that the characteristic structure of artifacts due to polar sampling should be similar in CT and MRI. Here, we considered random sampling patterns on a Cartesian grid, and therefore, the domain transfer method proposed in (30) is not directly applicable to our reconstructions that possess incoherent artifacts. Another recent, independent effort examined the reliability of reconstructions from a variational network to deviations in undersampling patterns and SNR between the training and testing domains (42). Mismatch in patterns or SNR between the two domains caused suboptimal performance even for modest acceleration factors. They also assessed the generalization capability by performing implicit domain transfer between PD-weighted knee images with and without fat suppression. A network trained on PD-weighted knee images without fat suppression was observed to yield relatively poor reconstructions of images with fat suppression and vice versa. Consistent with these observations, we also find that, without fine-tuning, networks trained on MR images of a given contrast (e.g. $T_1$-weighted) do not generalize well to images of a different contrast (e.g. $T_2$-weighted). That said, a distinct contribution of our work was to address the issue of data scarcity by training a network in a domain with ample data, and transferring the network to a domain with fewer samples.

An alternative approach proposed to train neural networks for MRI reconstruction with small datasets is Robust artificial-neural-networks for k-space interpolation, RAKI (13). This previous method aims to train a neural network for each individual subject that learns to synthesize missing k-space samples from acquired data. Unlike traditional k-space parallel imaging methods (50,60), a nonlinear interpolation kernel was estimated from central calibration data. Such nonlinear interpolation was shown to boost reconstruction performance beyond linear methods. However, RAKI might yield suboptimal performance when the optimal interpolation kernel shows considerable variation across k-space. Our proposed architecture for multi-coil reconstructions leverages a linear interpolation kernel, so the output of calibration-consistency blocks in our network can manifest similar reconstruction errors. Yet, the remaining CNN blocks are trained to recover fully-sampled reference images given images with residual artifacts at the output of CC blocks.

Here, we demonstrated domain transfer based on a cascade architecture with multiple CNNs interleaved with data- and calibration-consistency layers. The proposed approach might facilitate the use of neural networks for MRI reconstruction in applications where data are relatively scarce. It might also benefit other types of architectures that have been proposed for accelerated MRI (16,19,41), in particular architectures that require extensive datasets for adequate training (12,45). Here, the calibration-consistency projections were based on the SPIRiT method. These projections can also be replaced with other k-space methods for parallel imaging such as GRAPPA or RAKI. Note that the current study examined the generalization capability of networks trained on natural images to $T_1$-weighted and $T_2$-weighted images of the brain. ImageNet-trained networks could also be beneficial for reconstruction of MR images acquired with more specialized contrasts such as angiograms, and images acquired in other body parts.

# Tables

**Table 1.** Reconstruction quality for single-coil $T_1$-weighted images undersampled at R= 2, 4, 6, 8, 10. Reconstructions were performed via ImageNet-trained, $T_1$-trained and $T_2$-trained networks. PSNR and SSIM values are reported as mean±standard deviation across test images. Results are shown for raw networks trained on 2000 training images (raw), and fine-tuned networks tuned with few tens of $T_1$-weighted images (tuned).

|       |       | **ImageNet-trained** |              | **$T_1$-trained** |              | **$T_2$-trained** |              |
|-------|-------|------------------|--------------|------------------|--------------|------------------|--------------|
|       |       | *PSNR*           | *SSIM*       | *PSNR*           | *SSIM*       | *PSNR*           | *SSIM*       |
| **R=2** | **Raw**   | 40.33 ± 3.42 | 0.96 ± 0.02 | 40.65 ± 3.07 | 0.97 ± 0.01 | 40.15 ± 3.14 | 0.96 ± 0.01 |
|       | **Tuned** | 42.81 ± 3.32 | 0.97 ± 0.01 | 42.37 ± 3.25 | 0.97 ± 0.01 | 42.75 ± 3.22 | 0.97 ± 0.01 |
|       |       | **ImageNet-trained** |              | **$T_1$-trained** |              | **$T_2$-trained** |              |
|       |       | *PSNR*           | *SSIM*       | *PSNR*           | *SSIM*       | *PSNR*           | *SSIM*       |
| **R=4** | **Raw**   | 34.07 ±3.19  | 0.89 ± 0.03 | 34.87 ± 2.90 | 0.91 ± 0.02 | 33.26 ± 3.23 | 0.90 ± 0.03 |
|       | **Tuned** | 35.85 ± 3.03 | 0.93± 0.03  | 36.09 ± 3.19 | 0.93 ± 0.03 | 35.95 ±3.03  | 0.93 ± 0.03 |
|       |       | **ImageNet-trained** |              | **$T_1$-trained** |              | **$T_2$-trained** |              |
|       |       | *PSNR*           | *SSIM*       | *PSNR*           | *SSIM*       | *PSNR*           | *SSIM*       |
| **R=6** | **Raw**   | 29.42 ± 3.59 | 0.84 ± 0.04 | 32.34 ± 2.95 | 0.89 ± 0.03 | 30.48 ± 3.22 | 0.86 ± 0.03 |
|       | **Tuned** | 33.47 ± 3.11 | 0.90± 0.03  | 33.90 ± 3.26 | 0.90 ± 0.04 | 33.63 ± 3.09 | 0.90 ± 0.03 |
|       |       | **ImageNet-trained** |              | **$T_1$-trained** |              | **$T_2$-trained** |              |
|       |       | *PSNR*           | *SSIM*       | *PSNR*           | *SSIM*       | *PSNR*           | *SSIM*       |
| **R=8** | **Raw**   | 27.28 ± 3.77 | 0.81 ± 0.04 | 30.07 ± 3.18 | 0.86 ± 0.03 | 28.42 ± 3.14 | 0.83 ± 0.04 |
|       | **Tuned** | 32.14 ± 3.22 | 0.89 ± 0.04 | 32.21 ± 3.32 | 0.89 ± 0.04 | 32.17 ± 3.45 | 0.89 ± 0.04 |
|       |       | **ImageNet-trained** |              | **$T_1$-trained** |              | **$T_2$-trained** |              |
|       |       | *PSNR*           | *SSIM*       | *PSNR*           | *SSIM*       | *PSNR*           | *SSIM*       |
| **R=10** | **Raw**   | 25.82 ± 3.85 | 0.79 ± 0.05 | 28.84 ± 3.43 | 0.85 ± 0.04 | 27.72 ± 3.30 | 0.82 ± 0.04 |
|       | **Tuned** | 30.93 ± 3.40 | 0.87 ± 0.04 | 31.53 ± 3.38 | 0.88 ± 0.04 | 31.42 ± 3.28 | 0.88 ± 0.04 |

**Table 2.** Reconstruction quality for multi-coil $T_1$-weighted images undersampled at R= 2, 4, 6, 8, 10. Reconstructions were performed via ImageNet-trained and $T_1$-trained networks as well as SPIRiT. PSNR and SSIM values are reported as mean±standard deviation across test images. Results are shown for raw networks trained on 2000 training images (raw), and fine-tuned networks tuned with few tens of $T_1$-weighted images (tuned).

| | | ImageNet-trained | | $T_1$-trained | | SPIRiT | |
|---|---|---|---|---|---|---|---|
| R=2 | | *PSNR* | *SSIM* | *PSNR* | *SSIM* | *PSNR* | *SSIM* |
| | **Raw** | 49.48±1.73 | 0.995±.001 | 50.46±1.67 | 0.996±.002 | 50.47±1.68 | 0.996±.002 |
| | **Tuned** | 50.09±1.64 | 0.996±.002 | | | | |
| R=4 | | ImageNet-trained | | $T_1$-trained | | SPIRiT | |
| | | *PSNR* | *SSIM* | *PSNR* | *SSIM* | *PSNR* | *SSIM* |
| | **Raw** | 43.54±1.74 | 0.985±.006 | 45.36±1.75 | 0.989±.004 | 44.60±1.75 | 0.987±.004 |
| | **Tuned** | 44.76±1.76 | 0.989±.004 | | | | |
| R=6 | | ImageNet-trained | | $T_1$-trained | | SPIRiT | |
| | | *PSNR* | *SSIM* | *PSNR* | *SSIM* | *PSNR* | *SSIM* |
| | **Raw** | 39.55±1.90 | 0.970±.010 | 42.06±1.85 | 0.981±.006 | 40.62±1.73 | 0.975±.007 |
| | **Tuned** | 41.40±1.95 | 0.980±.007 | | | | |
| R=8 | | ImageNet-trained | | $T_1$-trained | | SPIRiT | |
| | | *PSNR* | *SSIM* | *PSNR* | *SSIM* | *PSNR* | *SSIM* |
| | **Raw** | 36.78±1.72 | 0.954±.014 | 39.14±1.75 | 0.971±.009 | 37.11±1.71 | 0.961±.012 |
| | **Tuned** | 38.45±1.77 | 0.969±.010 | | | | |
| R=10 | | ImageNet-trained | | $T_1$-trained | | SPIRiT | |
| | | *PSNR* | *SSIM* | *PSNR* | *SSIM* | *PSNR* | *SSIM* |
| | **Raw** | 34.27±1.72 | 0.943±.016 | 36.19±1.85 | 0.960±.012 | 34.23±1.72 | 0.948±.017 |
| | **Tuned** | 35.68±1.89 | 0.958±.014 | | | | |

# **Figures**

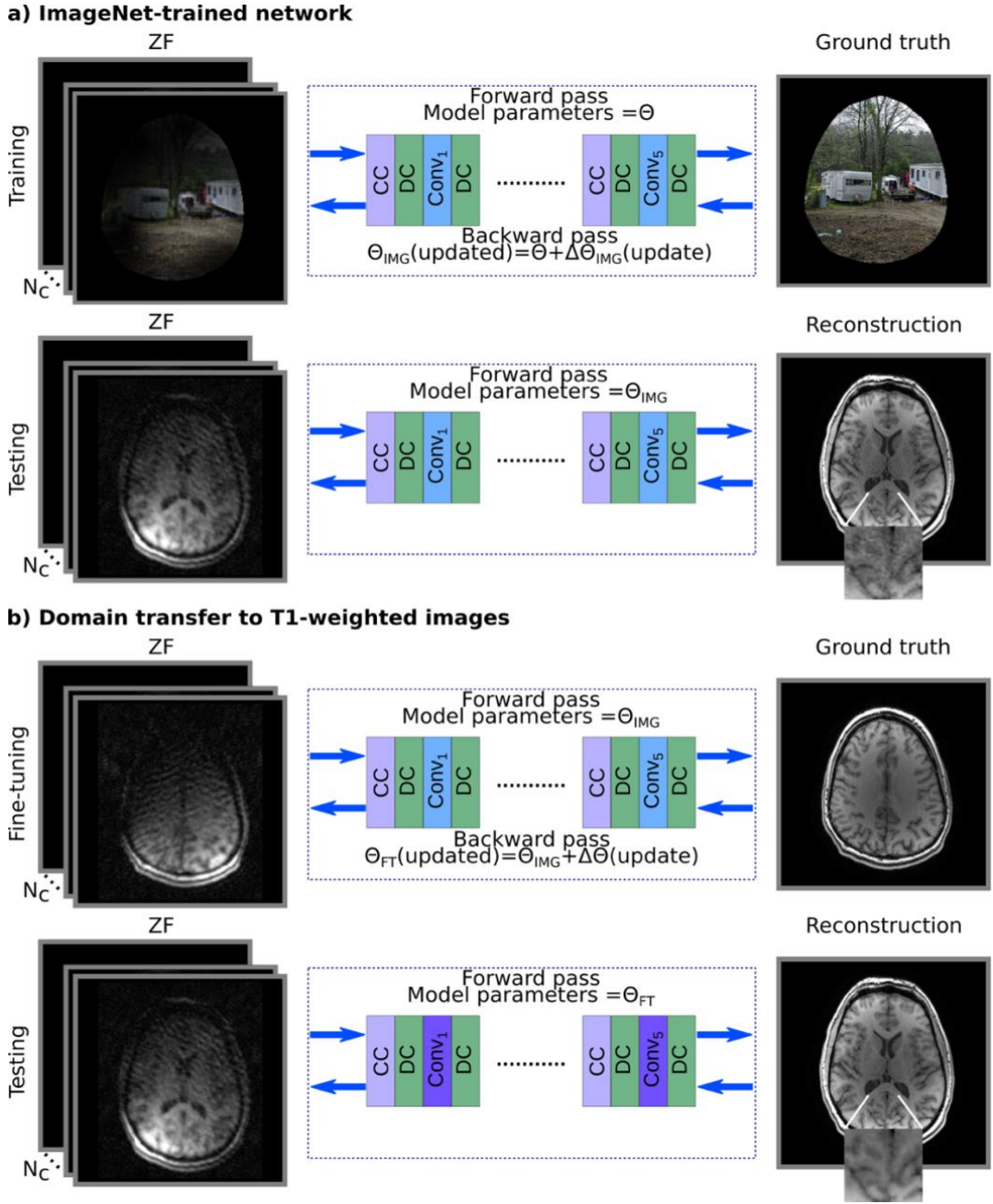

**Figure 1.** Proposed transfer-learning approach for NN-based reconstructions of multi-coil ($N_c$ coils) undersampled acquisitions. A deep architecture with multiple subnetworks is used. The subnetworks consist of calibration consistency "CC" and CNN "Conv" blocks, each followed by a data consistency block "DC". (a) Each subnetwork is trained sequentially to reconstruct synthetic multi-coil natural images from ImageNet, given zero-filled Fourier reconstructions of their undersampled versions. Due to differences in the characteristics of natural and MR images, the ImageNet-trained network will yield suboptimal performance when directly tested on MR images. (b) For domain transfer, the ImageNet-trained network is fine-tuned end-to-end in the testing domain using few tens of images. This approach enables successful domain transfer between natural and MR images.

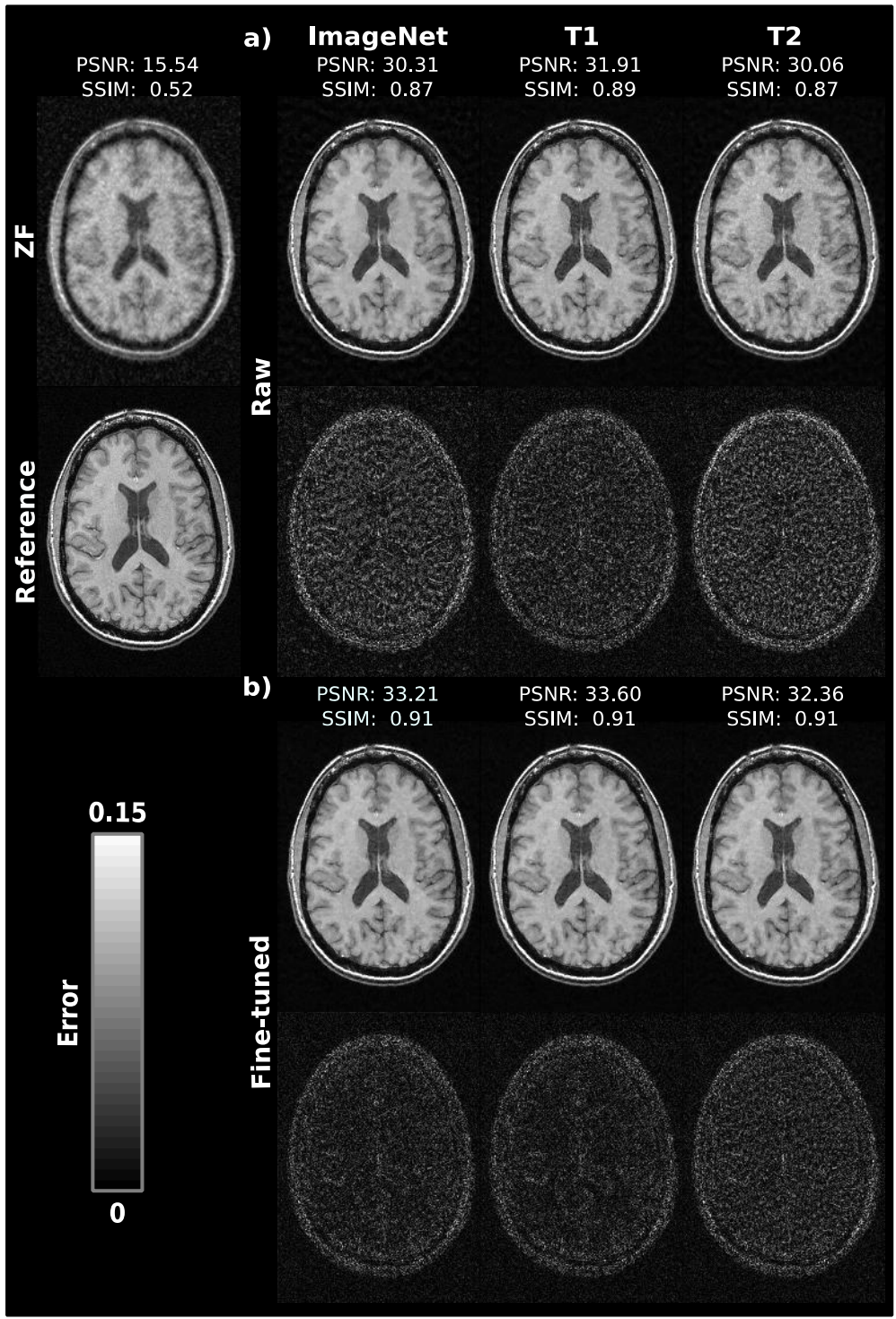

**Figure 2.** Representative reconstructions of a $T_1$-weighted acquisition at acceleration factor R=4. Reconstructions were performed via the Zero-filled Fourier method (ZF), and ImageNet-trained, $T_2$-trained, and $T_1$-trained networks. (a) Reconstructed images and error maps for raw networks (see colorbar). (b) Reconstructed images and error maps for fine-tuned networks. The fully-sampled reference image is also shown. Network training was performed on a training dataset of 2000 images and fine-tuned on a sample of 20 $T_1$-weighted images. Following fine-tuning with few tens of samples, ImageNet-trained and $T_2$-trained networks yield reconstructions of highly similar quality to the $T_1$-trained network.

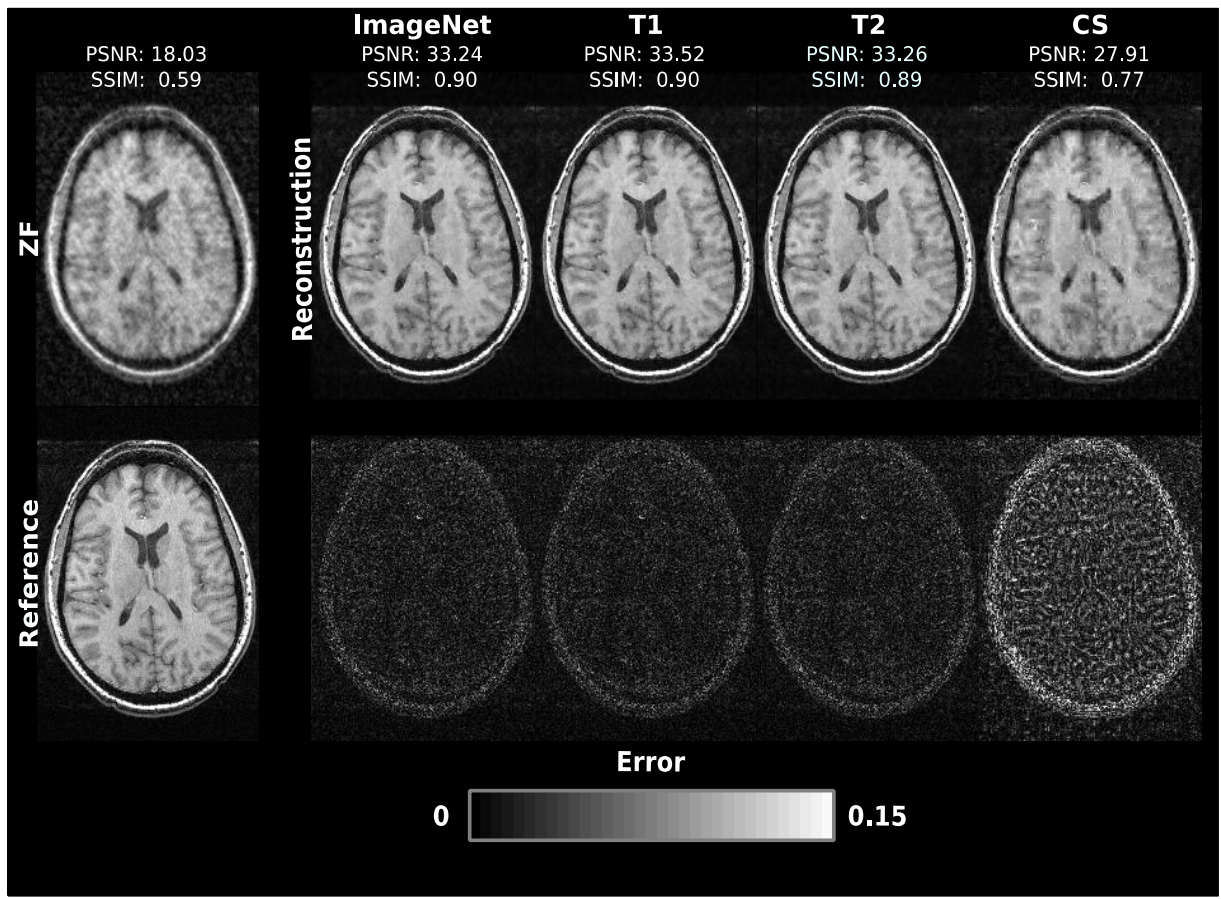

**Figure 3.** Reconstructions of a $T_1$-weighted acquisition with R=4 via ZF, conventional compressed-sensing (CS), and ImageNet-trained, $T_1$-trained and $T_2$-trained networks along with the fully-sampled reference image. Error maps for each reconstruction are shown below (see colorbar). Networks were trained on 2000 images and fine-tuned on 20 images acquired with the test contrast. The domain-transferred networks maintain nearly identical performance to the networks trained directly in the testing domain. Furthermore, the domain transferred-networks reconstructions outperform conventional CS in terms of image sharpness and residual aliasing artifacts.

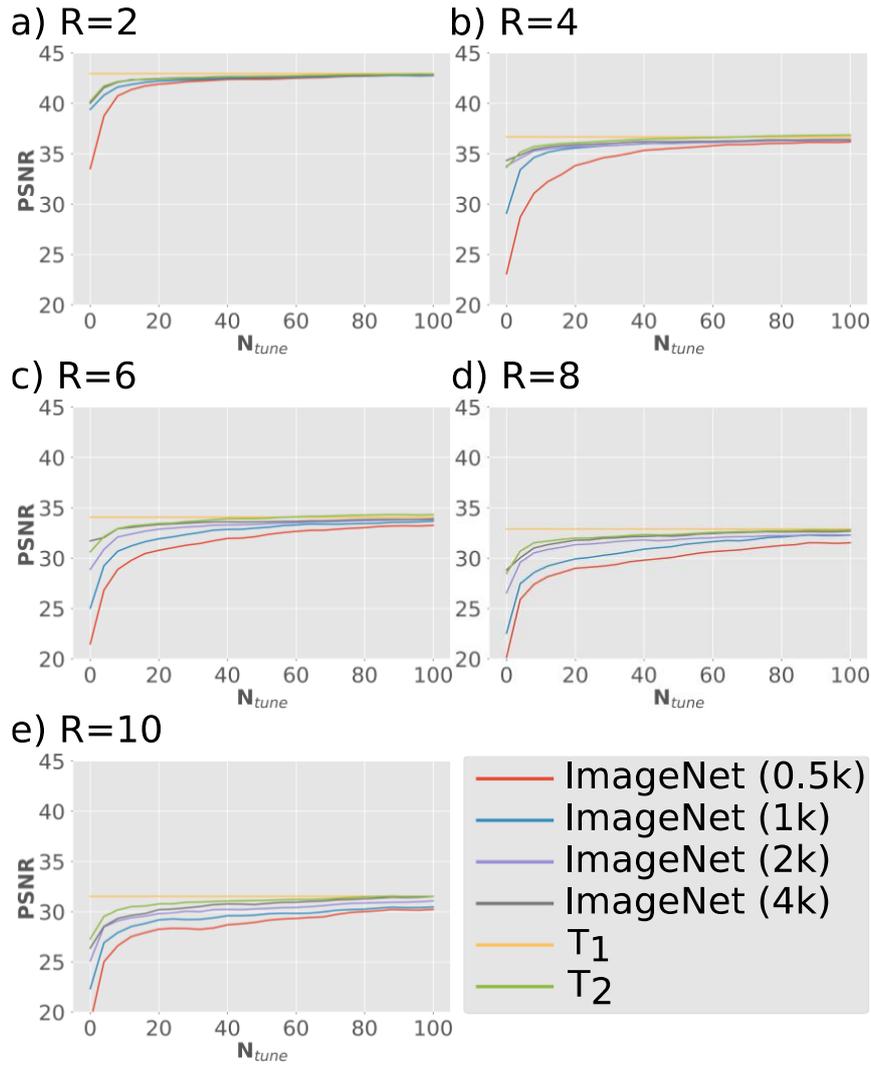

**Figure 4.** Reconstruction performance was evaluated for undersampled $T_1$-weighted acquisitions. Average PSNR values across $T_1$-weighted validation images were measured for the $T_1$-trained network (trained on 4k images and fine-tuned on 100 images), ImageNet-trained networks (trained on 500, 1000, 2000, or 4000 images), and $T_2$-trained network (trained on 4000 images). Results are plotted as a function of number of fine-tuning samples for acceleration factors (a) R=2, (b) R= 4, (c) R = 6, (d) R = 8, and (e) R= 10. Without fine-tuning, the $T_1$-trained network outperforms all domain-transferred networks. As the number of fine-tuning samples increases, the PSNR differences decay gradually to a negligible level. Domain-transferred networks trained on fewer samples require more fine-tuning samples to yield similar performance consistently across R.

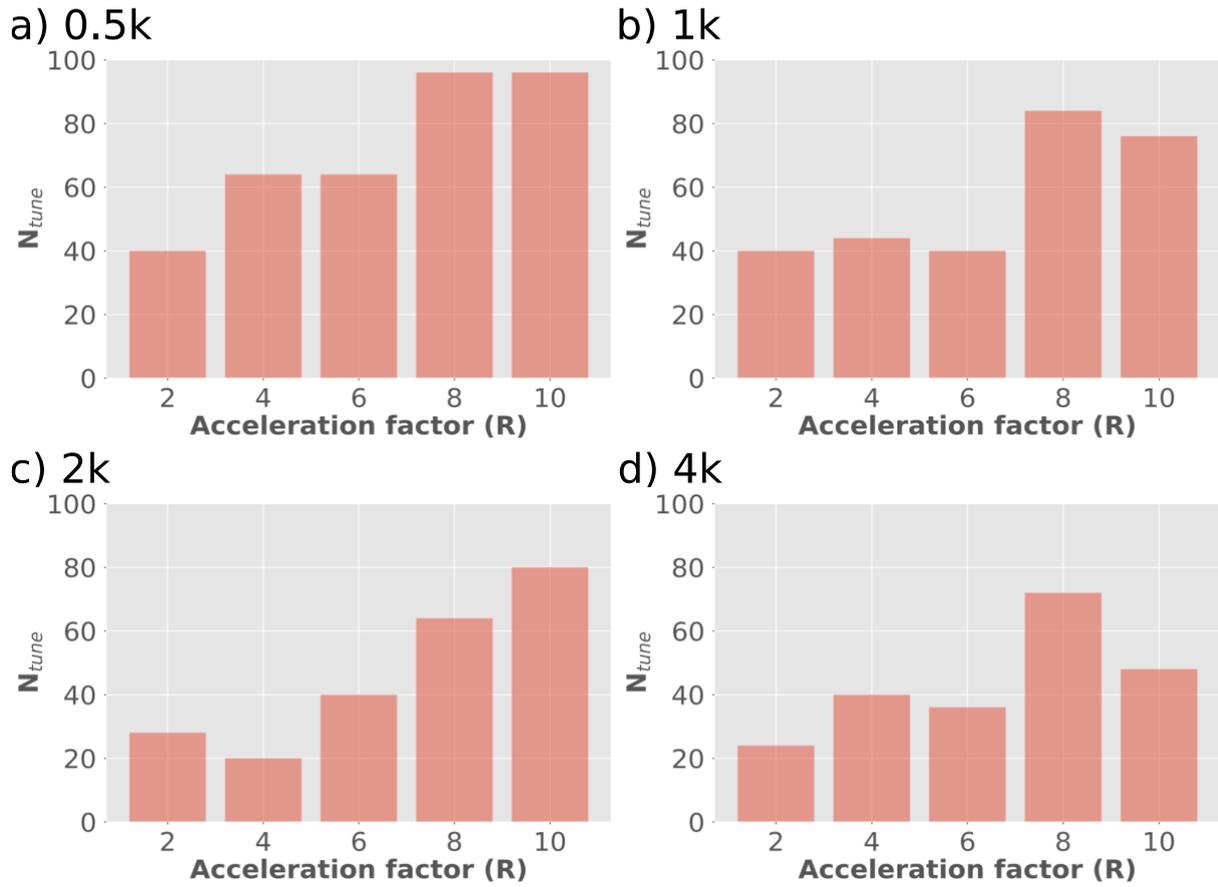

**Figure 5.** Number of fine-tuning samples required for the PSNR values for ImageNet-trained networks to converge. Average PSNR values across $T_1$-weighted validation images were measured for the ImageNet-trained networks trained on (a) 500, (b) 1000, (c) 2000, and (d) 4000 images. Convergence was taken as the number of fine-tuning samples where the percentage change in PSNR by incrementing $N_{tune}$ fell below 0.05% of the average PSNR for the $T_1$-trained network (see Figure 4). Domain-transferred networks trained on fewer samples require more fine-tuning samples for the PSNR values to converge. Furthermore, at higher values of R, more fine-tuning samples are required for convergence.

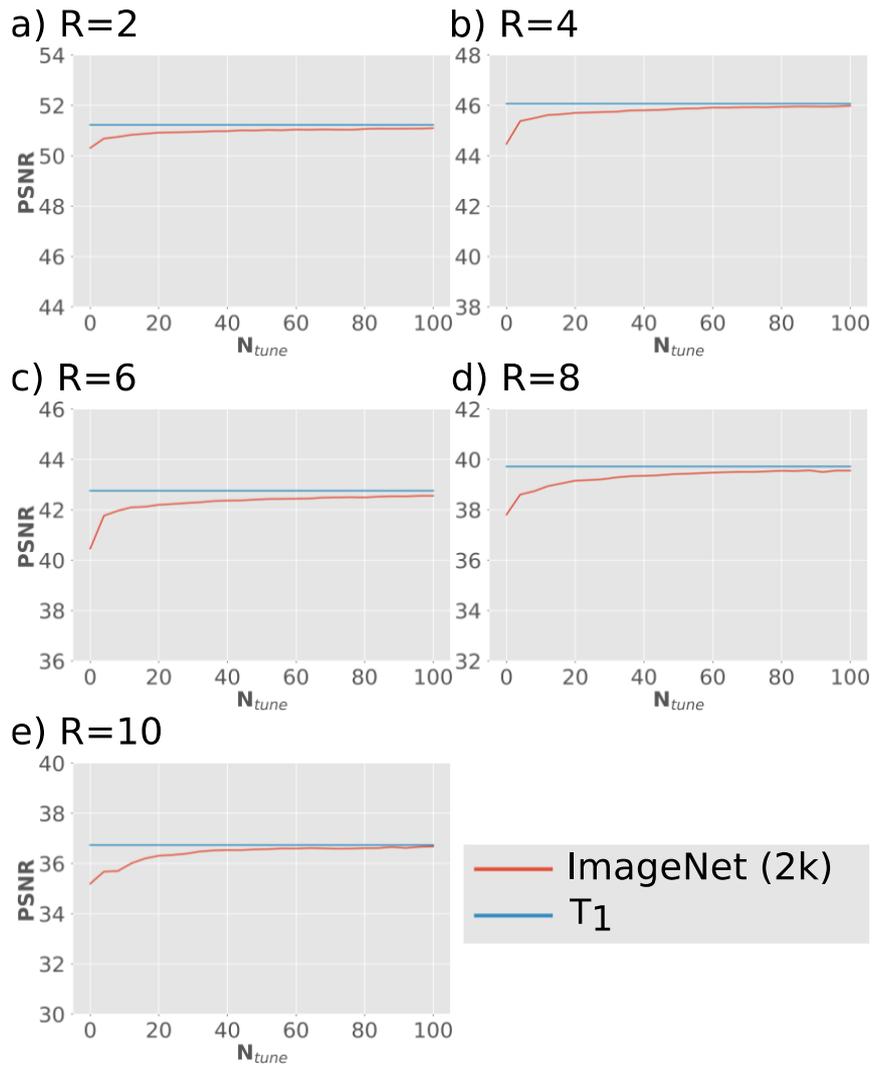

**Figure 6.** Reconstruction performance was evaluated for undersampled multi-coil $T_1$-weighted acquisitions. Average PSNR values across $T_1$-weighted validation images were measured for the $T_1$-trained network (trained and fine-tuned on 360 images), and ImageNet-trained network trained on 2000 images. Results are plotted as a function of number of fine-tuning samples for acceleration factors (a) R=2, (b) R= 4, (c) R = 6, (d) R = 8, and (e) R= 10. Without fine-tuning, the $T_1$-trained network outperforms the domain-transferred network. As the number of fine-tuning samples increases, the PSNR differences decay gradually to a negligible level.

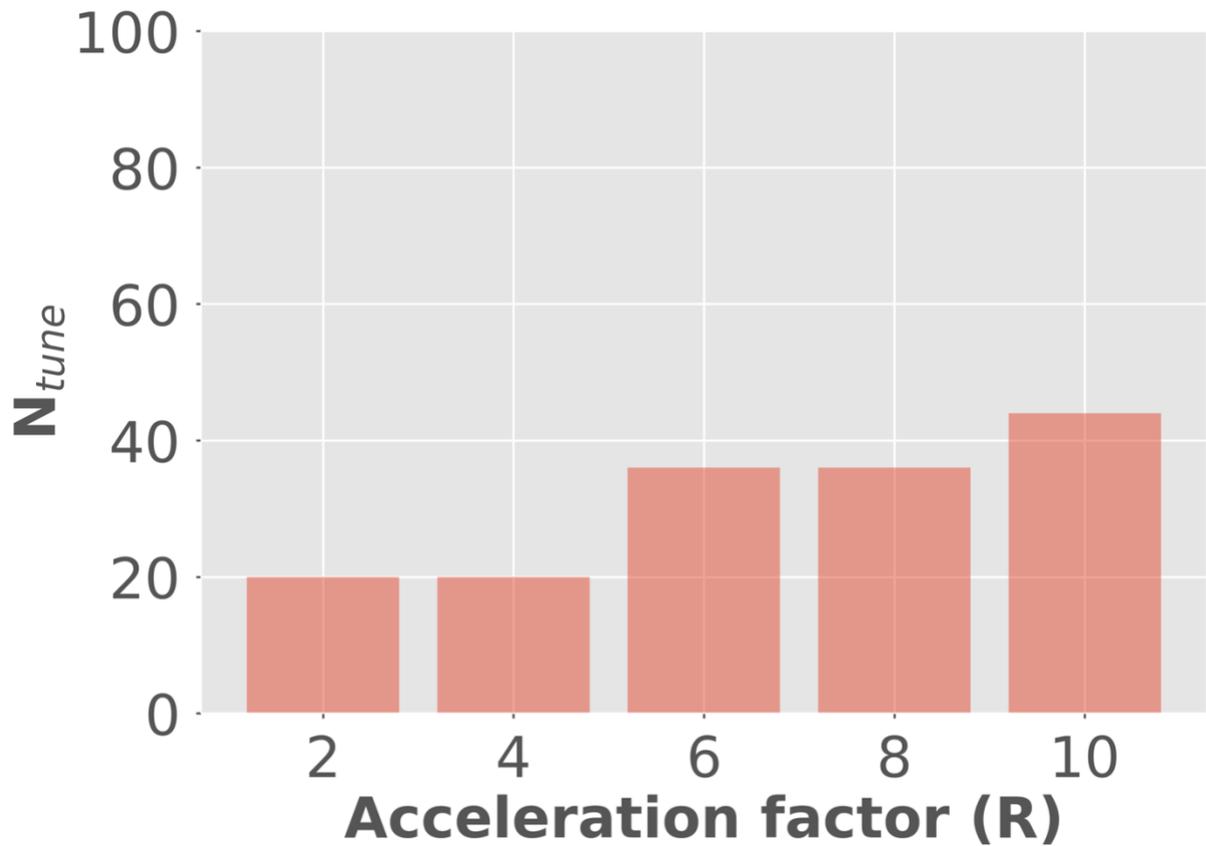

**Figure 7.** Number of fine-tuning samples required for the PSNR values for ImageNet-trained networks to converge. Average PSNR values across $T_1$-weighted validation images were measured for the ImageNet-trained network trained on 2000 images. Convergence was taken as the number of fine-tuning samples where the percentage change in PSNR by incrementing $N_{tune}$ fell below 0.05% of the average PSNR for the $T_1$-trained network (see Figure 6). At higher values of R, more fine-tuning samples are required for convergence.

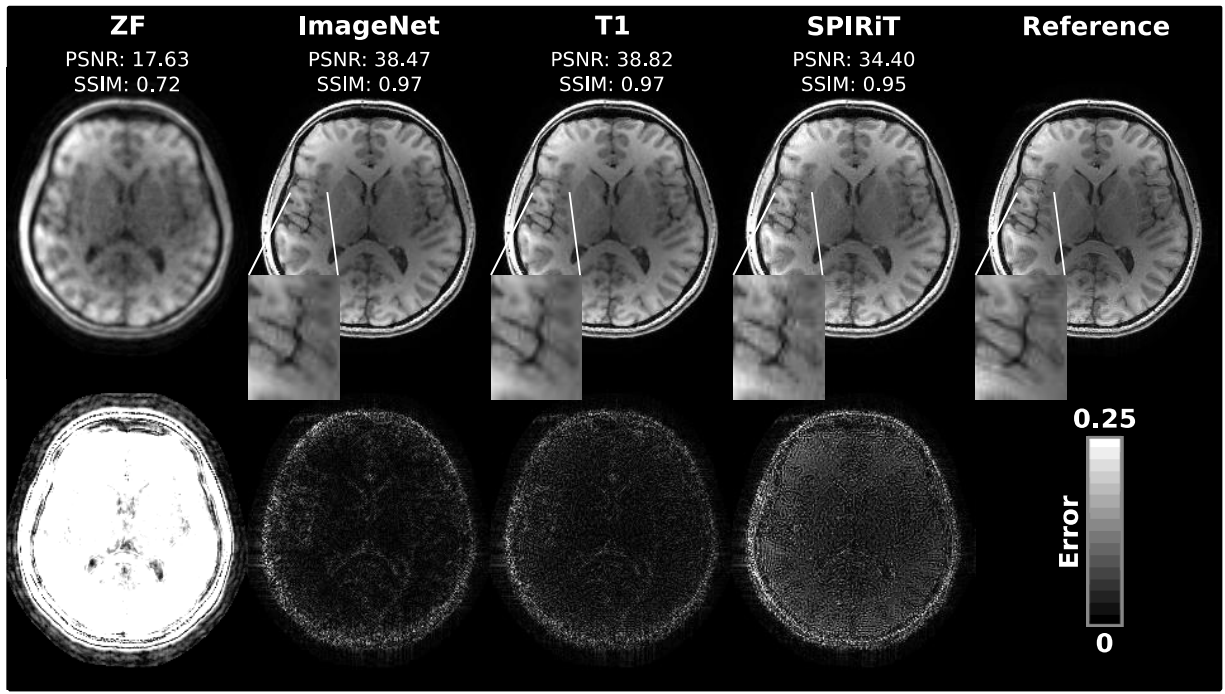

**Figure 8.** Representative reconstructions of a multi-coil $T_1$-weighted acquisition at acceleration factor R=10. Reconstructions were performed via ZF, ImageNet-trained and $T_1$-trained networks, and SPIRiT (top row). Corresponding error maps are also shown (see colorbar; bottom row) along with the fully-sampled reference (top row). Network training was performed on a training dataset of 2000 images and fine-tuned on a sample of 20 $T_1$-weighted images. The ImageNet-trained network maintains similar performance to the $T_1$-trained network trained directly on the images from the test domain. Furthermore, the domain-transferred network outperforms conventional SPIRiT in terms of residual aliasing artifacts.

# Supplementary Materials
# Table of Contents

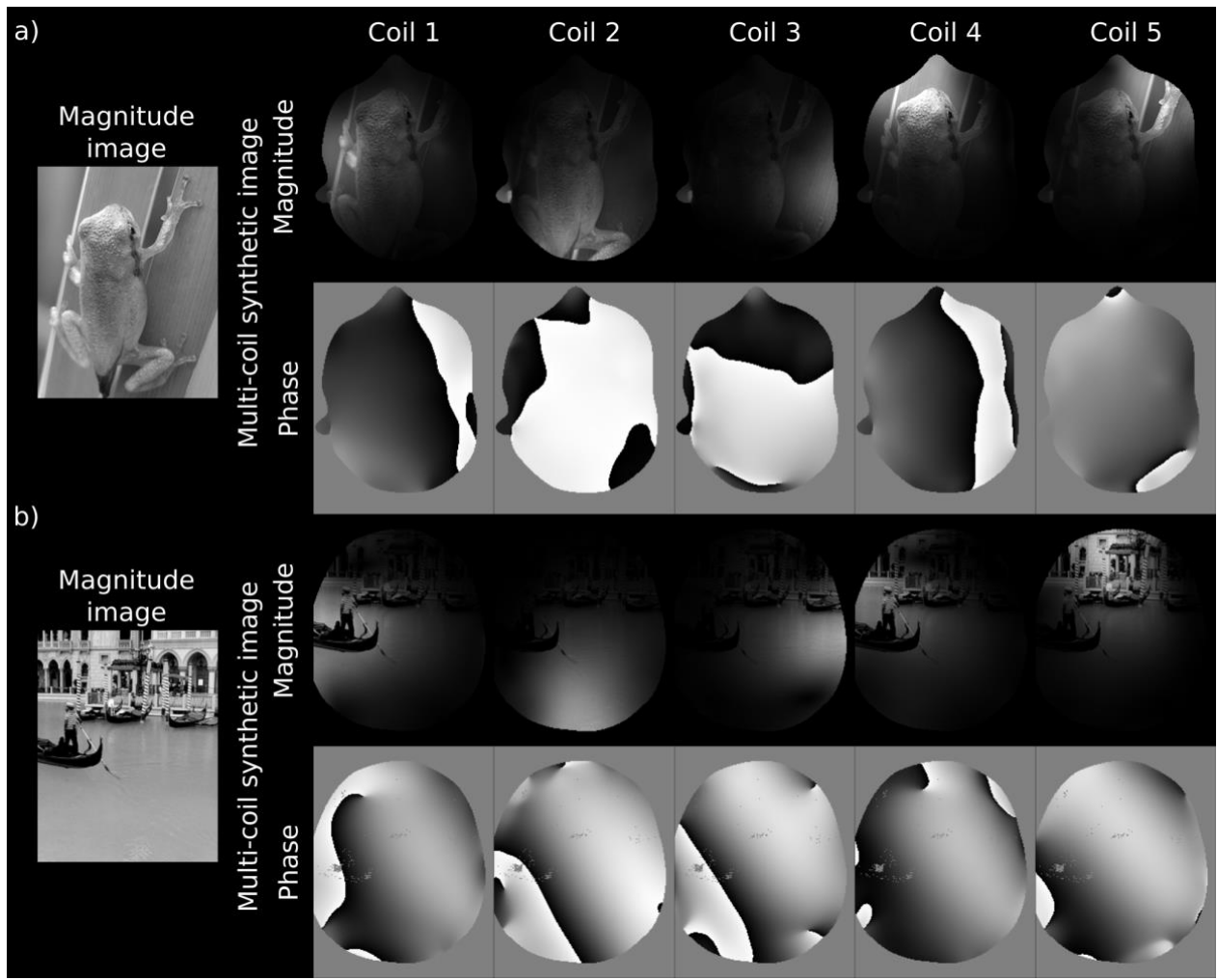

**Supplementary Figure 1.** Representative synthetic complex multi-coil natural images. Complex multi-coil natural images were simulated from magnitude images in ImageNet (see Methods for details). Magnitude and phase of two sets of simulated images (a and b) are shown along with their reference magnitude images.

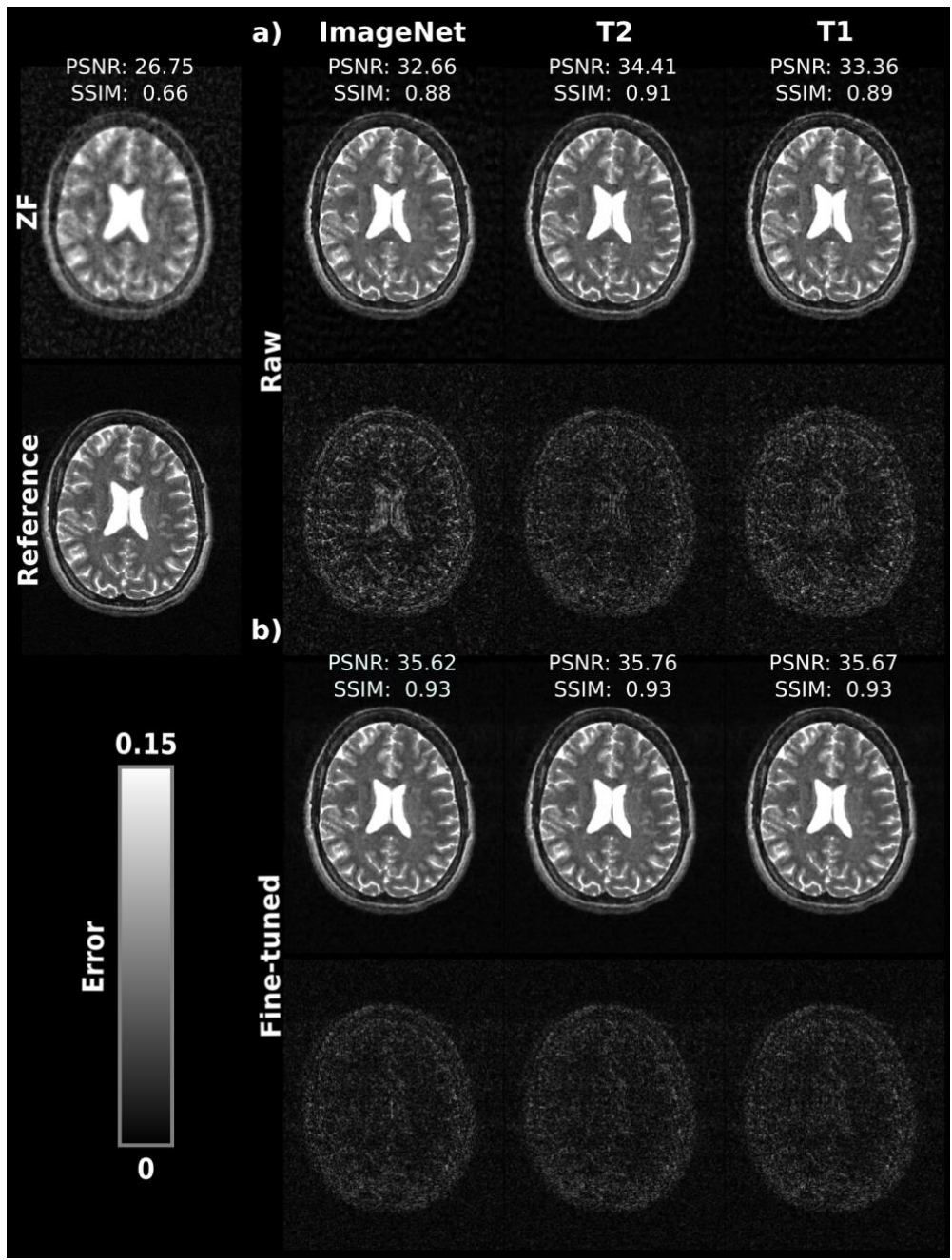

**Supplementary Figure 2.** Representative reconstructions of a $T_2$-weighted acquisition at acceleration factor R=4. Reconstructions were performed via the Zero-filled Fourier method (ZF), and ImageNet-trained, $T_2$-trained, and $T_1$-trained networks. (a) Reconstructed images and error maps for raw networks (see colorbar). (b) Reconstructed images and error maps for fine-tuned networks. The fully-sampled reference image is also shown. Network training was performed on a training dataset of 2000 images and fine-tuned on a sample of 20 $T_2$-weighted images. Following fine-tuning with few tens of samples, ImageNet-trained and $T_1$-trained networks yield reconstructions of highly similar quality to the $T_1$-trained network.

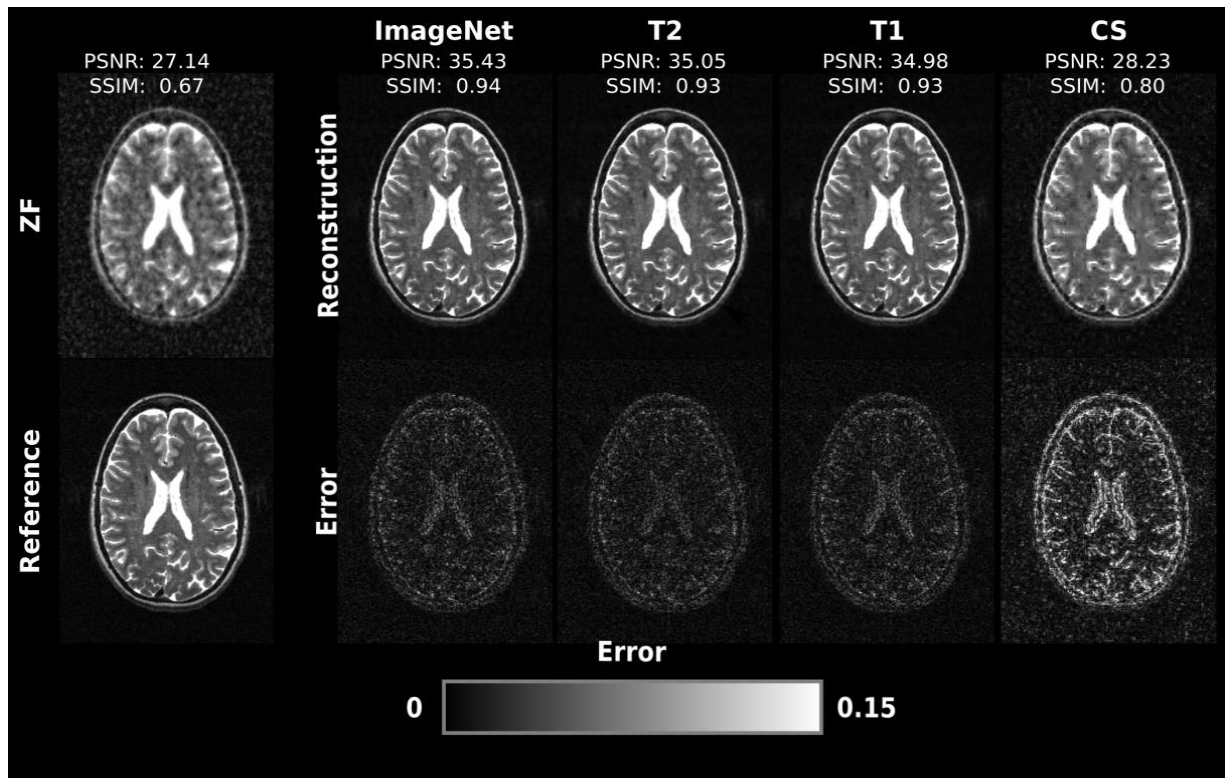

**Supplementary Figure 3.** Reconstructions of a T₂-weighted acquisition with R=4 via ZF, conventional compressed-sensing (CS), and ImageNet-trained, T₁-trained and T₂-trained networks along with the fully-sampled reference image. Error maps for each reconstruction are shown below (see colorbar). Networks were trained on 2000 images and fine-tuned on 20 images acquired with the test contrast. The domain-transferred networks maintain nearly identical performance to the networks trained directly in the testing domain. Furthermore, the domain transferred-networks reconstructions outperform conventional CS in terms of image sharpness and residual aliasing artifacts.

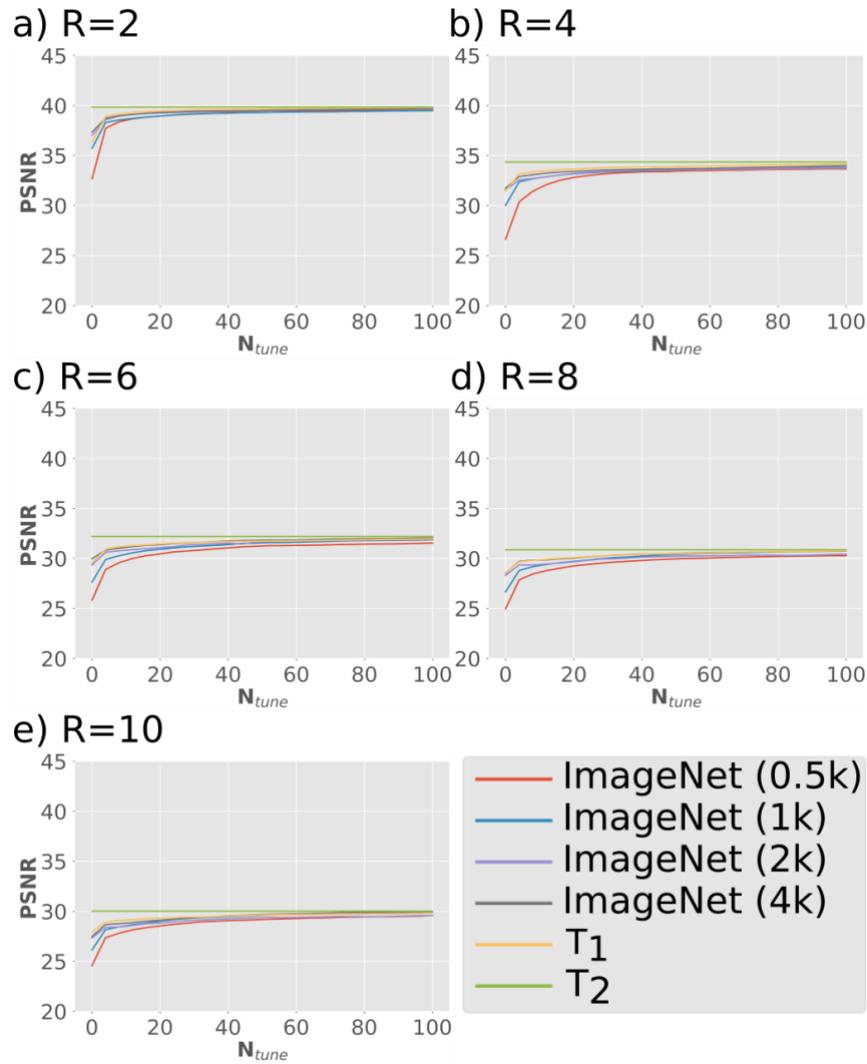

**Supplementary Figure 4.** Reconstruction performance was evaluated for undersampled $T_2$-weighted acquisitions. Average PSNR values across $T_2$-weighted validation images were measured for the $T_1$-trained network (trained on 4k images and fine-tuned on 100 images), ImageNet-trained networks (trained on 500, 1000, 2000, or 4000 images), and $T_1$-trained network (trained on 4000 images). Results are plotted as a function of number of fine-tuning samples for acceleration factors (a) R=2, (b) R= 4, (c) R = 6, (d) R = 8, and (e) R= 10. As the number of fine-tuning samples increases, the PSNR differences decay gradually to a negligible level. Domain-transferred networks trained on fewer samples require more fine-tuning samples to yield similar performance consistently across R.

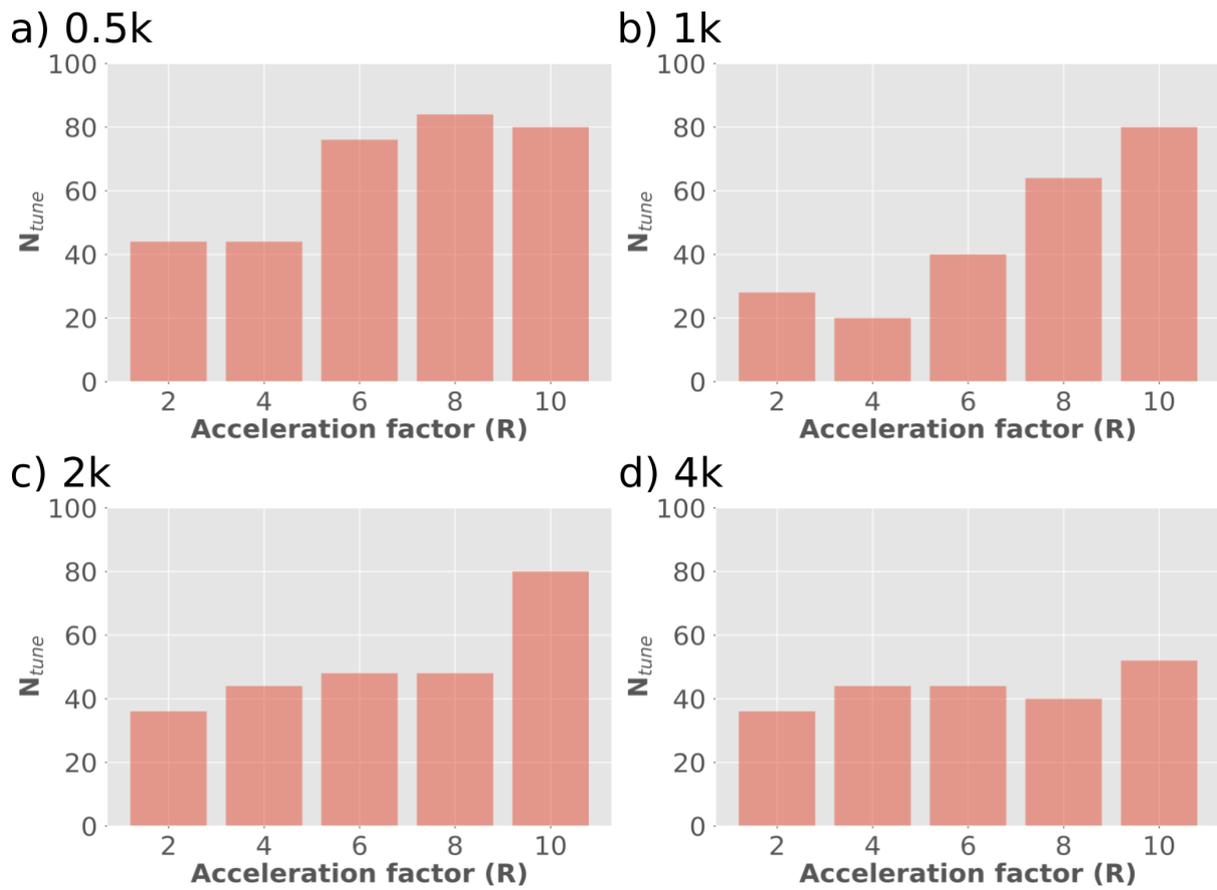

**Supplementary Figure 5.** Number of fine-tuning samples required for the PSNR values for ImageNet-trained networks to converge. Average PSNR values across T$_2$-weighted validation images were measured for the ImageNet-trained networks trained on (a) 500, (b) 1000, (c) 2000, and (d) 4000 images. Convergence was taken as the number of fine-tuning samples where the percentage change in PSNR by incrementing N$_{tune}$ fell below 0.05% of the average PSNR for the T$_2$-trained network (see Supp. Figure. 4). Domain-transferred networks trained on fewer samples require more fine-tuning samples for the PSNR values to converge. Furthermore, at higher values of R, more fine-tuning samples are required for convergence.

**Supplementary Table 1.** Reconstruction quality for single-coil $T_1$-weighted images undersampled at R= 2, 4, 6, 8, 10. Reconstructions were performed via ImageNet-trained, $T_1$-trained and $T_2$-trained networks, as well as conventional CS. PSNR and SSIM values are reported as mean±standard deviation across test images. Results are shown for raw networks trained on 2000 training images (raw), and fine-tuned networks tuned with few tens of $T_1$-weighted images (tuned).

| | | ImageNet-trained | | $T_1$-trained | | $T_2$-trained | |
|---|---|---|---|---|---|---|---|
| | | *PSNR* | *SSIM* | *PSNR* | *SSIM* | *PSNR* | *SSIM* |
| **R=2** | **Raw** | 40.33 ± 3.42 | 0.96 ± 0.02 | 40.65 ± 3.07 | 0.97 ± 0.01 | 40.15 ± 3.14 | 0.96 ± 0.01 |
| | **Tuned** | 42.81 ± 3.32 | 0.97 ± 0.01 | 42.37 ± 3.25 | 0.97 ± 0.01 | 42.75 ± 3.22 | 0.97 ± 0.01 |
| | **CS** | | | | | | |
| | | *PSNR* | | | *SSIM* | | |
| | | 37.54 ± 3.33 | | | 0.93 ± 0.24 | | |

| | | ImageNet-trained | | $T_1$-trained | | $T_2$-trained | |
|---|---|---|---|---|---|---|---|
| | | *PSNR* | *SSIM* | *PSNR* | *SSIM* | *PSNR* | *SSIM* |
| **R=4** | **Raw** | 34.07 ±3.19 | 0.89 ± 0.03 | 34.87 ± 2.90 | 0.91 ± 0.02 | 33.26 ± 3.23 | 0.90 ± 0.03 |
| | **Tuned** | 35.85 ± 3.03 | 0.93± 0.03 | 36.09 ± 3.19 | 0.93 ± 0.03 | 35.95 ±3.03 | 0.93 ± 0.03 |
| | **CS** | | | | | | |
| | | *PSNR* | | | *SSIM* | | |
| | | 31.77 ± 3.51 | | | 0.84 ± 0.04 | | |

| | | ImageNet-trained | | $T_1$-trained | | $T_2$-trained | |
|---|---|---|---|---|---|---|---|
| | | *PSNR* | *SSIM* | *PSNR* | *SSIM* | *PSNR* | *SSIM* |
| **R=6** | **Raw** | 29.42 ± 3.59 | 0.84 ± 0.04 | 32.34 ± 2.95 | 0.89 ± 0.03 | 30.48 ± 3.22 | 0.86 ± 0.03 |
| | **Tuned** | 33.47 ± 3.11 | 0.90± 0.03 | 33.90 ± 3.26 | 0.90 ± 0.04 | 33.63 ± 3.09 | 0.90 ± 0.03 |
| | **CS** | | | | | | |
| | | *PSNR* | | | *SSIM* | | |
| | | 29.71 ± 3.52 | | | 0.79 ± 0.05 | | |

| | | ImageNet-trained | | $T_1$-trained | | $T_2$-trained | |
|---|---|---|---|---|---|---|---|
| | | *PSNR* | *SSIM* | *PSNR* | *SSIM* | *PSNR* | *SSIM* |
| **R=8** | **Raw** | 27.28 ± 3.77 | 0.81 ± 0.04 | 30.07 ± 3.18 | 0.86 ± 0.03 | 28.42 ± 3.14 | 0.83 ± 0.04 |
| | **Tuned** | 32.14 ± 3.22 | 0.89 ± 0.04 | 32.21 ± 3.32 | 0.89 ± 0.04 | 32.17 ± 3.45 | 0.89 ± 0.04 |
| | **CS** | | | | | | |
| | | *PSNR* | | | *SSIM* | | |
| | | 28.56 ± 3.53 | | | 0.76 ± 0.06 | | |

| | | ImageNet-trained | | $T_1$-trained | | $T_2$-trained | |
|---|---|---|---|---|---|---|---|
| | | *PSNR* | *SSIM* | *PSNR* | *SSIM* | *PSNR* | *SSIM* |
| **R=10** | **Raw** | 25.82 ± 3.85 | 0.79 ± 0.05 | 28.84 ± 3.43 | 0.85 ± 0.04 | 27.72 ± 3.30 | 0.82 ± 0.04 |
| | **Tuned** | 30.93 ± 3.40 | 0.87 ± 0.04 | 31.53 ± 3.38 | 0.88 ± 0.04 | 31.42 ± 3.28 | 0.88 ± 0.04 |
| | **CS** | | | | | | |
| | | *PSNR* | | | *SSIM* | | |
| | | 27.98 ± 3.49 | | | 0.75 ± 0.06 | | |

**Supplementary Table 2.** Reconstruction quality for single-coil $T_2$-weighted images undersampled at R= 2, 4, 6, 8, 10. Reconstructions were performed via ImageNet-trained, $T_1$-trained and $T_2$-trained networks, as well as conventional CS. PSNR and SSIM values are reported as mean±standard deviation across test images. Results are shown for raw networks trained on 2000 training images (raw), and fine-tuned networks tuned with few tens of $T_2$-weighted images (tuned).

| | | ImageNet-trained | | $T_1$-trained | | $T_2$-trained | |
|---|---|---|---|---|---|---|---|
| | | PSNR | SSIM | PSNR | SSIM | PSNR | SSIM |
| R=2 | Raw | 38.84 ± 1.29 | 0.95 ± 0.01 | 38.30 ± 1.49 | 0.94 ± 0.01 | 39.93 ± 1.48 | 0.96 ± 0.01 |
| | Tuned | 41.81 ± 1.29 | 0.97 ± 0.01 | 41.38 ± 1.31 | 0.97 ± 0.01 | 41.79 ± 1.29 | 0.97 ± 0.01 |
| | CS | | | | | | |
| | | PSNR | | | SSIM | | |
| | | 35.94 ± 1.30 | | | 0.92 ± 0.01 | | |
| R=4 | | ImageNet-trained | | $T_1$-trained | | $T_2$-trained | |
| | | PSNR | SSIM | PSNR | SSIM | PSNR | SSIM |
| | Raw | 33.00 ± 1.46 | 0.88 ± 0.02 | 32.81 ± 1.60 | 0.87 ± 0.02 | 33.94 ± 1.51 | 0.90 ± 0.02 |
| | Tuned | 35.30 ± 1.38 | 0.92 ± 0.01 | 35.62 ± 1.40 | 0.93 ± 0.01 | 35.45 ± 1.41 | 0.92 ± 0.01 |
| | CS | | | | | | |
| | | PSNR | | | SSIM | | |
| | | 29.79 ± 1.51 | | | 0.81 ± 0.03 | | |
| R=6 | | ImageNet-trained | | $T_1$-trained | | $T_2$-trained | |
| | | PSNR | SSIM | PSNR | SSIM | PSNR | SSIM |
| | Raw | 30.35 ± 1.38 | 0.84 ± 0.02 | 30.90 ± 1.47 | 0.85 ± 0.03 | 31.68 ± 1.38 | 0.87 ± 0.02 |
| | Tuned | 33.05 ± 1.36 | 0.90 ± 0.02 | 32.98 ± 1.39 | 0.90 ± 0.02 | 33.15 ± 1.40 | 0.90 ± 0.02 |
| | CS | | | | | | |
| | | PSNR | | | SSIM | | |
| | | 27.71 ± 1.54 | | | 0.75 ± 0.03 | | |
| R=8 | | ImageNet-trained | | $T_1$-trained | | $T_2$-trained | |
| | | PSNR | SSIM | PSNR | SSIM | PSNR | SSIM |
| | Raw | 29.03 ± 1.40 | 0.81 ± 0.02 | 29.37 ± 1.44 | 0.82 ± 0.03 | 30.28 ± 1.37 | 0.85 ± 0.02 |
| | Tuned | 31.44 ± 1.37 | 0.87 ± 0.02 | 31.49 ± 1.39 | 0.88 ± 0.02 | 31.70 ± 1.36 | 0.88 ± 0.02 |
| | CS | | | | | | |
| | | PSNR | | | SSIM | | |
| | | 26.74 ± 1.55 | | | 0.72 ± 0.03 | | |
| R=10 | | ImageNet-trained | | $T_1$-trained | | $T_2$-trained | |
| | | PSNR | SSIM | PSNR | SSIM | PSNR | SSIM |
| | Raw | 27.96 ± 1.40 | 0.77 ± 0.03 | 28.29 ± 1.44 | 0.80 ± 0.03 | 29.51 ± 1.34 | 0.84 ± 0.03 |
| | Tuned | 30.64 ± 1.35 | 0.86 ± 0.02 | 31.01 ± 1.35 | 0.87 ± 0.02 | 31.10 ± 1.34 | 0.87 ± 0.02 |
| | CS | | | | | | |
| | | PSNR | | | SSIM | | |
| | | 26.16 ± 1.54 | | | 0.71 ± 0.04 | | |